\documentclass{article}
\usepackage{ijcai23}

\usepackage{fancyhdr} 
\usepackage{times}
\usepackage{soul}
\usepackage{url}
\usepackage[hidelinks]{hyperref}
\usepackage[utf8]{inputenc}
\usepackage[small]{caption}
\usepackage{graphicx}
\usepackage{amsmath}
\usepackage{amsthm}
\usepackage{booktabs}
\usepackage{algorithm}
\usepackage{algorithmic}
\usepackage{amssymb}
\usepackage[switch]{lineno}
\usepackage{xcolor}
\usepackage{multirow}
\usepackage{stfloats}
\usepackage{makecell}

\fancypagestyle{firstpage}{%
  \fancyhf{} 
  \fancyfoot[L]{Ongoing work} 
  \fancyfoot[C]{\thepage}     
}

\fancypagestyle{defaultstyle}{%
  \fancyhf{} 
  \fancyfoot[C]{\thepage}     
}

\newtheorem{definition}{Definition}

\title{Deep Learning within Tabular Data: Foundations, Challenges, Advances and Future Directions}

\author{
Weijieying Ren$^1$
\and
Tianxiang Zhao$^1$\and
Yuqing Huang$^{2}$\And
Vasant Honavar$^1$
\affiliations
$^1$Information Sciences and Technology, The Pennsylvania State University\\
$^2$Computer Science and Technology, University of Science and Technology of China\\
\emails
\{wjr5337, tkz5084, vuh14\}@psu.edu,
enthlinn@mail.ustc.edu.cn
}

\begin{document}

\maketitle
\thispagestyle{firstpage} 

\pagestyle{defaultstyle}

\section{Introduction}

\subsection{Related Surveys}
Based on the background discussed above, designing a state-of-the-art representation learning method for tabular data involves three fundamental elements: training data, network architectures, and learning objectives.
To improve both the quantity and quality of training data, various data-related techniques, such as data augmentation and generation, are employed or introduced.
To better leverage the inherent properties of tabular data, the neural architectures are specifically designed to capture: 1) irregular patterns within each column, including variations in scale, mean, and variance.
and 2) complex inter-relationships among different columns.
Finally, multiple learning strategies and objectives are defined to facilitate the learning of high-quality representations.

Despite incorporating three key design elements, most existing surveys on tabular data representation learning primarily concentrate on either neural architectural features or learning methodologies. 
An early survey article \cite{sahakyan2021explainable} provides a comprehensive overview of Explainable Artificial Intelligence (XAI) techniques applicable to tabular data, with a particular emphasis on feature transformation and classical machine learning methods in classification tasks.
Subsequently, several surveys have explored synthetic data generation methods. 
For instance, \cite{sauber2022use} reviews research on data generation within the healthcare domain, specifically focusing on GAN-based techniques. 
In addition, \cite{sauber2022use} investigates the application of GANs to address the class imbalance problem. 
Among these works, \cite{borisov2022deep} presents a comprehensive survey that discusses feature transformation, neural network design, and data generation challenges within a broader context.
In contrast, \cite{wang2024survey} systematically reviews and summarizes the recent advancements and challenges associated with self-supervised learning for non-sequential tabular data \cite{ren2024esacl}. 
Furthermore, with the emergence of foundation models and transformer-based large language models (LLMs), recent articles \cite{ruan2024language} examine the adaptation of these models to tabular data, concentrating primarily on their learning aspects.

Unlike previous works, we present a comprehensive review of representation learning methods for tabular data, focusing on their universality—effectiveness across a range of downstream tasks. Our discussion provides insights into the underlying intuitions driving these methods and examines how they enhance the quality of learned representations across all three key design aspects.
Specifically, we aim to identify and analyze research directions informed by recent state-of-the-art studies that concentrate on neural architecture design, the formulation of corresponding learning objectives, and the effective utilization of training data to improve the quality of learned representations for various downstream tasks. Table 1 highlights the distinctions between our survey and existing literature.

\subsection{Survey Scope and Literature Collection}
For literature review, we use the following keywords and inclusion criteria to collect literatures.

\textbf{Keywords}. ``tabular'' or ``'table'', ``tabular'' AND ``'representation'', ``tabular'' AND ``'embedding'', ``tabular'' AND ``'modeling'', ``transaction data'' AND ``'representation'', ``biomedical data'' AND ``'representation''.
We use these keywords to search well-known repositories, including ACM Digital Library, IEEE Xplore, Google Scholars, Semantic Scholars, and DBLP, for the relevant papers.

\textbf{Inclusion Criteria}
Related literatures found by the above keywords are further filtered by the following criterion. Only papers meeting these criteria are included for review.
\begin{itemize}
\item Written exclusively in the English language
\item Focused on approaches based on deep learning or neural networks
\item Published in or after 2020 in reputable conferences or high-impact journals
\end{itemize}

\textbf{Quantitative Summary}  Given the above keywords and inclusion criteria, we selected 127 papers in total. 
Fig. 3 shows the quantitative summary of the paper selected for review. We can notice from Fig. 3a that neural architectures and learning objectives are similarly considered important in designing state-of-the-art methods. Most papers were published at ICLR, NeurIPS, followed by AAAI and ICML (Fig. 3b). According to Fig. 3c, we expect more papers on this topic will be published in the future.

\section{Preliminary}
This section provides definitions and notations used in the paper, describes downstream tasks for tabular data analysis, and highlights the unique properties of tabular data.
\subsection{Definitions}
\begin{definition}
(Tabular Data). Tabular data is systematically arranged in a structured format characterized by rows and columns. Each row denotes an individual sample record, while each column signifies a distinct type of feature observation. Each column is composed of a header and a series of values, commonly referred to as cells.
Each row is represented as a vector of $M_{\text{num}}$ numerical features and $M_{\text{cat}}$ categorical features $\boldsymbol{x} = [\{\boldsymbol{x_{\text{num}}^i}\}_{i=1}^{M_{\text{num}}},\{\boldsymbol{x_{\text{cat}}^i}\}_{i=1}^{M_{\text{cat}}}]$, where $\boldsymbol{x^i_{\text{num}}} \in R$ and $\boldsymbol{x^i_{\text{cat}}} \in \{1,2,...,C_i\}$.
$C_i$ denotes the size of finite candidate values for the $i$-th categorical
feature.
\end{definition}


\begin{definition}
(Classification).
Tabular data classification aims to assign predefined class labels 
$Y = \{y_1, y_2,...,y_C \}$ to each row of tabular data.
Denoted $D$ as a tabular dataset with $N$ samples, $X$ is a row of tabular data and $y$ is the corresponding labels.
where $y_i \in \{-1,1\}$ is a binary classification task and $y_i \in \{1,2,...,C\}$ is a multi-class classification task. 
 
\end{definition}

\begin{definition}
(Regression).
Tabular data regression has a similar goal to classification tasks, with a key difference in label annotation. In tabular data regression, the objective is to predict a continuous value $y \in R$ for each row. 

\end{definition}

\begin{definition}
(Clustering). Tabular data clustering aims to partition X into a group of clusters $G = {g_1, g_2,..., g_G}$ by maximizing the similarities between tabular rows within the same cluster
and the dissimilarities between tabular rows of different clusters. 
.
\end{definition}

\begin{definition}
(Anomaly Detection). Tabular anomaly detection is a crucial process for identifying tabular samples within a dataset that significantly deviate from established patterns of normal behavior. 
This approach typically involves training a model on a labeled dataset $D$, where the model learns the characteristics that define normal observations. Once trained, the model computes anomaly scores $A = (a_i,...,a_{|X|})$ for each row in an unseen test set $X_{test}$.
These scores quantify the degree of deviation for each observation. 
The final classification of anomalies is made by comparing each score $a_i$ against a predetermined threshold $\delta$: an sample is classified as anomalous if $a_i > \delta$ and as normal otherwise.
\end{definition}

\begin{definition}
(Imputation of Missing Values). Tabular imputation aims to fill missing values with plausible values to facilitate subsequent analysis. 
Given tabular data $X$ and known binary matrix $M \in R$, $x_i$ is missing if $m_i = 0$, and is observed otherwise. 
The imputed tabular data is given as:
$X_{imputed} = X \odot M + \hat{X} \odot (1-M)$.
\end{definition}

\begin{definition}
(Retrieval). Tabular retrieval aims to obtain a set of samples that are
most similar to a query provided. Given a query sample $X$ and a similarity measure $f(\cdot)$,
find an ordered list $Q = \{X_{i=1}^k\}$
of tabular samples in the given dataset or database, containing tabular samples that are the most similar to tabular queries.
\end{definition}

\begin{definition}
(Test Time Adaptation on Tabular Data). Given a pre-trained source model $f(\cdot)$, TTA adapts source domain model parameters $\theta$ to obtain target domain parameters $\theta'$ using unlabeled
target domain data $DT = {(xi)}NT i=1$. 
It should be noted that the feature spaces for the source and target domains are identical; however their distributions are not.
\end{definition}

\subsection{Foundational Properties In Tabular data Modeling}
In this subsection, we elaborate on the distinctive characteristics of tabular data and corresponding distinct research perspectives.
Due to these specific properties, methodologies developed for image or text data often unsuitable to be applied directly to tabular data.


\textbf{Heterogeneity} 
Tabular data exhibits heterogeneity due to the incorporation of various data types, including discrete entities such as categorical and binary features, continuous variables represented by numerical data. Furthermore, columns with the same data type may still display distinct marginal distributions, as evidenced by differences in statistical properties such as mean, variance, and scale.

\textbf{Diverse and Contextualized Semantic Meaning}
Tabular data exhibits diverse semantic meanings among its columns. 
For instance, in a clinical diagnosis prediction task, the value 100 is ambiguous and lacks meaning until contextualized, such as by specifying 100 kg or 100 ml.
In addition, the marginal distribution of features can vary across different contextual settings, even if they share the same semantic meaning. For example, in a weather prediction task, the statistical characteristics of air moisture may differ significantly between eastern and western regions.
This property complicates the transfer of knowledge across domains and tasks, and it also poses challenges for tabular imputation, often requiring the expertise of domain specialists.

\textbf{Permutation Invariance and Equivalence}
Permutation Invariance refers to the property that the outcomes of an analysis or model remain unchanged when the rows or columns of a tabular dataset are permuted. This means that rearranging the order of the observations (rows) or the features (columns) does not affect the statistical properties, relationships, or predictions derived from the data.
Permutation Equivalence indicates the tabular dataset remains essentially the same in terms of its statistical properties or relationships, regardless of the order of observations or features. 
Consequently, the results of analyses, such as predictions or statistical measures, remain consistent even when specific operations, such as normalization or scaling of features, are applied.

\textbf{High Noise and Missing Value}
Tabular data, especially in real-world environments, often contains noise and missing values. 
This noise typically arises from measurement errors or annotation mistakes. Missing values can also stem from measurement errors and may exhibit random missing patterns.
In some cases, missing values can convey important information and indicate "missing not at random" patterns. For example, in electronic health record data, each patient may undergo only specific lab tests that are critical for accurate diagnosis verification. These partial observations actually provide essential information for a model to understand the relationships between biomedical features and their corresponding diagnosis labels.

\textbf{Inter-Relationship across Columns or Rows}
In tabular data, both columns and rows reveal associations.
To model column-wise interactions, existing works analyze this property mainly from two directions:
(1) \textbf{Statistical Dependency} Different columns may have relevance with each other due to overlapped semantic information. For example, `age' and `date of birth' share statistical overlap when recording patient demographic information. Consequently, it is desired to exploit such dependency during encoding and modeling tabular data distribution,
(2) \textbf{Latent Relations}  There would be latent causal factors behind multiple column observations. With this observation, related columns or abstract concepts can be integrated and processed collectively to extract such critical latent factors. For instance, the family history of disease exhibits a distinct semantic correspondence to diabetes. Hence, the critical factor behind the user's vulnerability can be reflected by collecting and aggregating columns regarding family disease histories.
Meanwhile, summarizing group-level behaviors and modeling the corresponding relationships among samples are essential for enhancing model prediction accuracy. For instance, various medical settings are characterized by the presence of multiple distinct patient behaviors. Identifying and characterizing these subgroups is crucial for understanding underlying diseases and improving the delivery of medical care.

Particularly, researches have been made against heterogeneity the following dimensions:
(1) Modeling with different distributions for encoding columns of different data formats or frequencies of elements.
(2) Separately learning on different columns with metrics that are sound w.r.t information theory. The basic assumption is that disparities exist across information contained in each columns.
(3) Designing special element-level representation units for each column. For example in a continuous column, its numerical distribution range can be partitioned into distinct bins, with each data value being transformed into the index of corresponding interval bin. This transformation can map the raw input into a more interpretable and robust representation space.

\subsection{Applications}

\textbf{HealthCare}
Healthcare records employ tabular data structures for the storage and management of comprehensive patient health information \cite{ma2022elucidating,ma2022learning,ma2023learning}. Each column corresponds to specific data types, encompassing personal details, medical images \cite{liu2017left,li2019lumen}, medical history, and diagnostic and treatment information. The utilization of tabular data in healthcare tasks leads to various classifications: 
\begin{itemize}
    \item \textbf{Patient Outcome Prediction}. Examples include patient mortality prediction , predicting the diagnosis of diseases, and forecasting patient responses to specific drugs.
    \item \textbf{Clinical Trial Outcome Prediction}. This involves predicting the likelihood of a clinical trial succeeding in obtaining approval for commercialization.
    \item  \textbf{Tabular Search and Retrieval Problems}. Tasks under this category encompass clinical trial retrieval, where relevant trials are identified based on a given query or input trial; and insurance retrieval, involving the search for patient historical information.
    \item \textbf{Tabular Generation}. An example is Trial Patient Simulation, where the generation of synthetic clinical trial patient records facilitates data sharing across institutes while safeguarding patient privacy.
    \item \textbf{Tabular Data Transference} Tabular data transfer is fundamental to Health Information Exchange initiatives, where different healthcare entities share patient data securely. This promotes coordinated care, supports multidisciplinary care, reduces duplicate tests, and enhances overall healthcare efficiency.
\end{itemize}


\textbf{E-commerce}
In e-commerce, tabular data is widely used for various purposes to organize, analyze, and present information related to products, transactions, and customer interactions. 
The industrial practice of tabular data can be summarized into the following categories: 
\begin{itemize}
    \item \textbf{Product Recommendations}. E-commerce platforms use tabular data to generate personalized product recommendations. Columns include customer inquiries, support tickets, resolutions, customer browsing and purchase history. 
    \item \textbf{Transaction Fraud Detection}. Tabular data is utilized for monitoring transactions and detecting fraudulent activities. Columns may include transaction details, payment methods, and fraud indicators.
    \item \textbf{Product Search}. E-commerce platforms employ tabular search and retrieval for product searches. Users can search for products based on various criteria, such as category, price range, and specifications, and the system retrieves matching products.
    \item  \textbf{Tabular Transference}. Transfer learning allows e-commerce businesses to leverage existing knowledge from one context to improve performance in related tasks, reducing the need for extensive training on new datasets. It can lead to more efficient model training, faster adaptation to new markets Real-world application relates to Supply Chain Optimization, Dynamic Pricing prediction and Customer Lifetime Value Prediction.
\end{itemize}

\textbf{Energy Management}
Energy Management: Utilities and energy companies use tabular data to monitor energy consumption, analyze usage patterns, and optimize distribution networks for efficient energy management. Realworld application relates to \textbf{peak demand prediction}, \textbf{Renewable Energy Source Prediction}, \textbf{Carbon Emission Data Retrieval}.


\subsection{Foundational Properties In Tabular data Modeling}

Learning from tabular data poses a significant challenge due to the heterogeneous column contents. Different columns usually contain distinct semantics and display diverse distributions, leading to difficulties in learning the representation space. Furthermore, latent relations often exist behind observed columns, and modeling such interactions could be important for discovering critical factors.  Extensive researches have been conducted in this domain, and in this section we elaborate on these properties and distinct research perspectives on top of them.

\textbf{Heterogeneous} Tabular data exhibits heterogeneity due to the inclusion of diverse data types, encompassing discrete entities such as categorical and binary features, as well as continuous variables like numerical features.
Meanwhile, columns sharing identical data types may still manifest distinct marginal distributions, exemplified by variations in statistical properties such as mean and variance even when both following the Gaussian distribution.
Particularly, researches have been made against heterogeneity the following dimensions:
(1) Modeling with different distributions for encoding columns of different data formats or frequencies of elements.
(2) Separately learning on different columns with metrics that are sound w.r.t information theory. The basic assumption is that disparities exist across information contained in each columns.
(3) Designing special element-level representation units for each column. For example in a continuous column, its numerical distribution range can be partitioned into distinct bins, with each data value being transformed into the index of corresponding interval bin. This transformation can map the raw input into a more interpretable and robust representation space.

\textbf{Interaction} In tabular data, columns (or abstract concepts) frequently have relations with other.
Essentially, existing works analyze this property mainly from two directions:
\begin{itemize}
    \item \textbf{Statistical Dependency} Different columns may have relevance with each other due to overlapped semantic information. Consequently, it is desired to exploit such dependency during encoding and modeling tabular data distribution.
    \item \textbf{Collective Encoding}  There would be latent causal factors behind multiple column observations. With this observation, related columns or abstract concepts can be integrated and processed collectively to extract such critical latent factors. For instance, the family history of disease exhibits a distinct semantic correspondence to diabetes. Hence, the critical factor behind the user's vulnerability can be reflected by collecting and aggregating columns regarding family disease histories.
\end{itemize}

Explorations against column-wise interaction can be categorized based on the type of assumed interaction information and the interaction mechanism. Consideration points include the extent and dimensions in which information can be shared across columns (feature-level) and rows (sample-level), and the form of statistical association such as linear or non-linear formulations. Interaction modeling is also important for the missing value problem. In many realistic tasks, tabular data often suffers from incompleteness due to data privacy or mistakes in data collection , including biostatistics \cite{mackinnon2010use}, 
finance , 
algriture , etc. Column-wise interaction modeling is critical for the quality of imputation against missing values.

\subsection{Applications}

\textbf{HealthCare}
Healthcare records employ tabular data structures for the storage and management of comprehensive patient health information. Each column corresponds to specific data types, encompassing personal details, medical history, and diagnostic and treatment information. The utilization of tabular data in healthcare tasks leads to various classifications: 
(I) \textbf{Patient Outcome Prediction}. Examples include patient mortality prediction, predicting the diagnosis of diseases \cite{liang2024inducing}, forecasting patient responses to specific drugs and to depression \cite{qin2023read,zhao2023skill}.
(II) \textbf{Clinical Trial Outcome Prediction}. This involves predicting the likelihood of a clinical trial succeeding in obtaining approval for commercialization.
(III) \textbf{Tabular Search and Retrieval Problems}. Tasks under this category encompass clinical trial retrieval, where relevant trials are identified based on a given query or input trial; and insurance retrieval, involving the search for patient historical information.
(IV) \textbf{Tabular Generation}. An example is Trial Patient Simulation, where the generation of synthetic clinical trial patient records facilitates data sharing across institutes while safeguarding patient privacy.
(V) \textbf{Tabular Data Transference} Tabular data transfer is fundamental to Health Information Exchange initiatives, where different healthcare entities share patient data securely. This promotes coordinated care, supports multidisciplinary care, reduces duplicate tests, and enhances overall healthcare efficiency.

\textbf{E-commerce}
In e-commerce, tabular data is widely used for various purposes to organize, analyze, and present information related to products, transactions, and customer interactions. 
The industrial practice of tabular data can be summarized into the following categories: 
(I) \textbf{Product Recommendations} . E-commerce platforms use tabular data to generate personalized product recommendations. Columns include customer inquiries, support tickets, resolutions, customer browsing and purchase history. 
(II) \textbf{Transaction Fraud Detection}. Tabular data is utilized for monitoring transactions and detecting fraudulent activities. Columns may include transaction details, payment methods, and fraud indicators.
(III) \textbf{Product Search}. E-commerce platforms employ tabular search and retrieval for product searches. Users can search for products based on various criteria, such as category, price range, and specifications, and the system retrieves matching products.
(IV) \textbf{tabular Transference}. Transfer learning allows e-commerce businesses to leverage existing knowledge from one context to improve performance in related tasks, reducing the need for extensive training on new datasets. It can lead to more efficient model training, faster adaptation to new markets Real-world application relates to Supply Chain Optimization , Dynamic Pricing prediction  and Customer Lifetime Value Prediction .


\textbf{Energy Management}
Energy Management: Utilities and energy companies use tabular data to monitor energy consumption, analyze usage patterns, and optimize distribution networks for efficient energy management. Realworld application relates to \textbf{peak demand prediction} , \textbf{Renewable Energy Source Prediction} , \textbf{Carbon Emission Data Retrieval} .

\section{Tabular Data Modeling}
In this section, we present an elaborated taxonomy of three aspects in tabular data learning: \textbf{Attributes Representation, Inter-Column Dependency Modeling, Side Learning Tasks}. 
We deliver a thorough analysis, delving into their primary challenges and typical strategies.

\subsection{Heterogeneous Attributes Encoding}

\textbf{Challenges}  Tabular learning aims to capture the intrinsic characteristics of each attribute (column). However, given that different tabular columns often originate from diverse sources and exhibit a high degree of heterogeneity, adopting a uniform encoding strategy becomes suboptimal. Current research predominantly focuses on three key dimensions within this domain:

\textbf{Sub-Challenge 1: Diverse Data Formats and Distributions} Heterogeneous tabular data 
frequently encompasses attribute formats including real-valued variables, categorical tags, textual descriptions, etc. Attributes within the same format may also display distinct distributions and ranges. This diversity necessitates the design of attribute-specific encoding strategies—for example, encoding real-valued variables using learned Gaussian models and mapping categorical tags to vector spaces. 
As a result, learning a homogeneous representation space for all attributes poses a challenge but is crucial for facilitating model training and designing effective loss functions.

\textbf{Sub-Challenge 2: Attributes Representation Spaces}. Data is inherently composed of elementary units representing the fundamental components that may not be further decomposed. Choices of elementary units involve trade-offs between representation ability, interpretability, and robustness. For instance, an image of a bird can be depicted as a collection of pixels or a composition of bird parts. 
Consequently, tabular representation learning also revolves around identifying the basic elements within tabular columns to devise effective representation strategies.

\textbf{Sub-Challenge 3:  Incorporating Semantic Domain Knowledge}. The magnitude of a cell in isolation may lack meaningful interpretation without contextual information from the tabular header. For instance, the value "100" can convey distinct meanings depending on whether it is associated with 'PH' or 'Heart Rate'. Enhancing tabular representation, generalization, and robustness involves exploring, harnessing, and integrating rich contextual semantics from headers into tabular cells. This integration is crucial for ensuring the meaningful interpretation of tabular data in various contexts.

\textbf{Solution to Sub-Challenge 1.} Addressing the challenge of diverse data, it is proposed to conduct homogeneous learning and transform the raw mixed-type attributes into a unified space with continuous properties, which can also ease the gradient-based learning pipeline. A popular strategy is``feature tokenizer," that converts each numerical and categorical column into a $d$-dimensional vector, e.g., a linear transformation \cite{huang2020tabtransformer,gorishniy2021revisiting}. 
Another strategy involves a two-stage training process where the first stage aims to generate a more homogeneous representation of the data across dimensions, subsequently utilized by the second stage. Typically, after the generative model, e.g., VAE has been effectively trained, the latent embeddings are extracted through the encoder and serve as the input for the subsequent stage.

\textbf{Solution to Sub-Challenge 2.} In addition to previous studies that focused on converting raw tabular data, especially categorical information, into a continuous space, the second group of research concentrates on the fundamental elements of tabular representation learning. Taking inspiration from the widely used one-hot encoding algorithm for categorical features discretizes numerical features into intervals, replacing original values with discrete descriptors. These descriptors  can be represented using piecewise linear encoding or periodic activation functions.
Another approach involves representing each cell as a new form of "column header is value." A concatenation of these cells is considered a fundamental representation of each row record.

\textbf{Solution to Sub-Challenge 3}.
The key insight in this direction is that there is often abundant, auxiliary domain information that can further describing input feature semantic representations and interactions \cite{nguyen2019mtab,santos2022knowledge}. 
Research works in this direction aims to enhance semantic representation through the usage of external knowledge like knowledge graphs \cite{liu2023tabular}.
The learning procedure can be summarized as the following steps: 
(1) \textbf{Knowledge-graph construction} they construst an auxiliary KG to describe input feature, in which each input feature corresponds to a node in the
auxiliary KG. 
(2) \textbf{Node Embedding}. Each input feature $j$ associated to a learnable weight vector $\theta_j \in \mathbb{R}^h$, e.g., MLP, such that the weight vectors of all d features compose the weight matrix $\theta \in \mathbb{R}^{d \times h}$.
(3) \textbf{Feature interaction estimation.}
A trainable message-passing function could be learned to further update node embedding, based on the assumption that two input features which correspond to similar nodes in the KG should have similar weight vectors in the node embedding space \cite{ruiz2023high}.

Notably, tabular semantic representation is a prevalent topic in Natural Language Processing \cite{arik2021tabnet}, e.g., Tabular QA and Tabular semantic parsing. However, the majority of benchmark datasets in this domain enriches full-text descriptions, which falls outside the scope of this survey. Interested readers are directed to \cite{} for a comprehensive exploration.

\subsection{Inter-Column Dependency Modeling}
In tables, different columns may exhibit overlapping semantics and correlations, with decision-making often contingent upon latent factors behind certain interactions. For example, in a clinical setting, correlations exist behind patient demographic attributes such as gender, age and vulnerability to certain diseases. The patient's response to a specific medical diagnosis may also be intricately linked to their familial history of inherited diseases. Formulating data dependencies behind tabular columns improves representation learning and facilitates downstream decision-making, representing a central challenge.

In tables, different columns may exhibit overlapping semantics and correlations, with decision-making often contingent upon latent factors behind certain interactions. For example, in a clinical setting, correlations exist behind patient demographic attributes such as gender, age and vulnerability to certain diseases. The patient's response to a specific medical diagnosis may also be intricately linked to their familial history of inherited diseases. Formulating data dependencies behind tabular columns improves representation learning and facilitates downstream decision-making, representing a central challenge.
\textbf{Challenges} In tabular data, different attributes work complementary to each other and intricate relationships often exist behind their observed values. Capturing these dependency structures is crucial for modeling the observed data and enhancing representation capability. Moreover, deep neural networks (DNNs) encounter challenges when learning from dense numerical tabular features due to the complexity of optimization hyperplanes in fully connected models, which increases the risk of converging to local optima \cite{fernandez2014we}. Structure modeling can alleviate this challenge by uncovering critical factors and ease the learning burden. Commonly employed data structures include  trees \cite{silva2020optimization,zhao2022towards}, graphs \cite{zhou2022table2graph,zhao2024interpretable}, and capsules \cite{chen2022tabcaps} and logic rules \cite{rentablog}. For example, tree-based method shows its merit in iteratively picking the features with the largest statistical information gain \cite{chen2016xgboost}. 
In this survey, we explore recent advancements to capture relations across tabular columns in three streads: 

\textbf{Sub-Challenge 1: Hierarchical Structure Modeling}.
Hierarchical Structure defines an arrangement where abstract concepts are delineated as a function of less abstract ones. This hierarchical relation is prevalent in tables. For example, a table may have columns representing country, region, and city, creating a hierarchy where cities are nested within regions, and regions are nested within countries. 
Identifying, capturing, and modeling latent hierarchical structures constitutes a fundamental challenge in this context.

\textbf{Sub-Challenge 2: Interactive Structures Modeling}.
In tables, different columns may exhibit overlapping semantics and correlations, with decision-making often contingent upon latent factors behind certain interactions. For example, in a clinical setting, correlations exist behind patient demographic attributes such as gender, age and vulnerability to certain diseases. The patient's response to a specific medical diagnosis may also be intricately linked to their familial history of inherited diseases. Formulating data dependencies behind tabular columns improves representation learning and facilitates downstream decision-making, representing a central challenge.

\textbf{Sub-Challenge 3: Latent Structure Discovery}.
The challenges presented by Challenge 1 and Challenge 2 raise a fundamental question regarding the modeling of data dependency structures. 
However, such structure is usually latent and unknown, particularly in situations where domain-specific knowledge is absent.
The discovery of such structures is essential for advancing structural modeling and serves as valuable evidence for root cause analysis, enhancing the explanatory capabilities of the model in a data-driven way.

\textbf{Solution to Sub-Challenge 1:}
Depending on the target of specific structure, such
methods can be divided into three sub-directions.
\textbf{Sub-Solution 1a: tree structure modeling}. DeepGBM \cite{ke2019deepgbm} is a pioneering work that integrates the advantages of DNN and GBDT. DeepGBM comprises two distinct neural network components: CatNN, specialized in managing sparse categorical features, and GBDT2NN, tailored for distilling knowledge from GBDT to process dense numerical features.
The NODE architecture \cite{popov2019neural} generalizes ensembles of oblivious decision trees, harnessing the advantages of both end-to-end gradient-based optimization and the efficacy of multi-layer hierarchical representation learning.
GrowNet \cite{badirli2020gradient} employs shallow neural networks as weak learners within a versatile gradient boosting framework. 
\cite{good2023feature} introduce an innovative framework that pioneers the alternation between sparse feature learning and differentiable decision tree construction. This approach aims to generate small, interpretable trees while maintaining high performance.

\textbf{Sub-Solution 1 b: neural module-based structure learning}. TABCAPs \cite{chen2022tabcaps} utilize capsule networks to model feature-wise interactions. 
In the primary capsule layer, each sample undergoes encoding into multiple vectorial features using optimizable multivariate Gaussian kernels \cite{}. Subsequently, a successive iterative process of feature clustering is applied to attain higher-level semantics. 
TANGOS \cite{jeffares2022tangos} utilizes a sparse and orthogonal regularization on the neural network, encouraging latent neurons to emphasize sparse, non-overlapping input features. This approach results in a collection of diverse and specialized latent units.
Net-DNF \cite{katzir2020net} presents an innovative framework characterized by an inherent inductive bias that yields models structured according to logical Boolean formulas in disjunctive normal form (DNF) over affine soft-threshold decision terms. Moreover, Net-DNFs actively promote localized decision-making processes executed over discrete subsets of the input features.

\textbf{Sub-Solution 1c: graph structure modeling}, with a considerable body of research dedicated to this direction. 
These approaches typically construct a graph based on predefined node and edge categories before using (heterogeneous variants of) graph neural networks to capture a representation of structure. 
Among these approaches, graph neural networks (GNN) are commonly employed to model
(i) Feature-wise Interactions that Captures interactions and dependencies between individual features;
(ii) Instance-wise Relations that models relationships between different instances or rows of the tabular data;
and (iii) Feature-Instance correlations that address correlations between features and specific instances within the table.
\textit{Feature-wise graph modeling} aims to automatically estimate and represent relations among tabular features in the form of a learnable weighted graph \cite{zhou2022table2graph,yan2023t2g}.
The learning procedure can be summarized into the following steps:
(1) \textbf{Graph Construction}. They represent each feature as a node and estimate pair-wise feature interactions as an adjacency matrix.
(2) \textbf{Graph Structure Learning}.
To estimate expressive feature relations and learn a reliable graph structure, prior works primarily focus on learning the adjacency matrix, considering three key factors:
1) node semantic meaning; 
2) implicit feature relations \cite{zhao2022exploring};
3) homogeneous assumption in GNN that heterogeneous tabular property violates the assumption of GNNs that connected nodes should exhibit similar patterns \cite{zhu2020beyond}.
\cite{yan2023t2g} generates data-adaptive edge weights by performing node semantic matching.
\cite{zhou2022table2graph} both consider node semantic matching and high-order interaction and achieve this by an ensemble graph that consists of an adaptive probability adjacency matrix with self attention and a static graph which calculate relation topology score through node semantic embeddings. 
Recently, \cite{wan2023graphfade} find that features from multiple fields in tabular data exhibit different patterns and it is difficult for all of them to be matched with one single graph. 
To address this, they first perform a hierarchically clustering method to separates initial tabular data into several parts with minimal correlation and then derived graph structures from decorrelated clusters that have minimal feature correlations.
(3) \textbf{Graph Sparfication}. To remove redundant feature relations and improve training efficiency without a constructed dense graph, they selectively collects salient features for the final prediction \cite{}. \cite{yan2023t2g} design a global readout node to selectively collect salient features from each layer. \cite{zhou2022table2graph} employ the reinforcement learning method to strengthen the key feature interaction connections.
\textit{Instance-wise graph modeling} treats each instance as a node in a graph and cast tabular classification as node classification problem \cite{liao2023tabgsl}. 
The third line of \textit{feature-instance graph modeling} aim to model instance-feature associates and perform message Propagation to enhance the target data instance representations. They construct a data-feature bipartite graph with data instance nodes and feature value nodes and then perform message passing on hypergraph neural network \cite{du2022learning}.

\textbf{Solution to Sub-Challenge 2}. Interactive Structures Modeling
formulates data dependency in a sequential decision-making framework, where the high-level concept depends on the local decisions made at every step.
\textbf{Sub-solution 2a} comprises transformer-based methods that leverage sequential attention mechanisms to determine which features to consider at each decision step. 
This approach enhances interpretability and improves learning efficiency by allocating learning capacity to the most salient features.
For instance, TabNet \cite{arik2021tabnet} designs a sequential attention mechanism for both feature selection and reasoning. 
SAINT \cite{somepalli2022saint} introduces the Self-Attention and Intersample Attention Transformer, allowing attention mechanisms over both rows and columns.
NAN \cite{luo2020network} introduces a Network On Network (NON) architecture tailored for tabular data classification. NON comprises three integral components: a field-wise network at the base, designed to capture intra-field information; an across-field network in the middle, dynamically selecting suitable operations based on data characteristics; and an operation fusion network at the summit, facilitating deep fusion of outputs from the chosen operations.

\textbf{Sub-Solution 2b} investigated using Differentiable Decision Tree (DDTs) for reinforcement learning tasks, with the goal to combine the flexibility of neural networks with the interpretable structure of decision trees. Existing works focus on policy learning using DDTs \cite{silva2020optimization,tambwekar2023natural,pace2022poetree}, on the contrary, \cite{kalra2023can} aims to learn interpretable reward functions using DDTs.

\textbf{Solution to Sub-Challenge 3}. Latent Structure Discovery aims to learn latent dependencies in a fully data-driven manner. Defining the structure of tabular data can be challenging, especially without domain knowledge. As an alternative approach, recent efforts have explored the use of differentiable strategies to autonomously search for the tabular structure, as seen in AGEBO-Tabular \cite{egele2021agebo} and TabNAS \cite{yang2022tabnas}. Another direction involves leveraging generative models for discovering latent structures, which will be elaborated upon in Section 3.3.

\subsection{Specialized Learning Tasks on Tabular Data}
In addition to strategies that enhance data representations by attribute-specific encoding and  relation modeling, a myriad of works focuses on specific tasks and learning goals on tabular data. These tasks are critical for the broader application of tables, and may provide additional learning signals to improve tabular representations. In this section, we introduce two representative side learning tasks, generative modeling of tabular data, and knowledge transfer across tables in similar domains.

\subsubsection{Tabular Data Generation Tasks}
Tabular data is a prevalent data format in various industrial sectors, including finance and electronic health records. However, the limited privacy considerations constrain the feasibility of releasing such data. 
Moreover, the collection of such data involves intricate procedures, such as unbalanced data nature, ensuring participant willingness and synchronized updates. Consequently, collected tabular datasets often contain missing values. 
The generation of authentic tabular data is crucial, particularly in the contexts of tabular data imputation and synthetic data creation. 
Existing works primarily devise model architectures based on popular generative models, including GAN \cite{goodfellow2020generative}, VAE \cite{kingma2013auto}, and the diffusion model \cite{wijmans1995solution}. These efforts aim to tackle three core challenges prevalent in tabular data generation:
1) Input Heterogeneity;
2) Sampling Quality;
3) Latent Structure Modeling.

As a pionering work in the realm of Generative Adversarial Networks (GANs), medGAN, as described by \cite{choi2017generating}, integrates both an auto-encoder and a GAN to effectively generate diverse sets of continuous and/or binary medical data. Subsequently, TableGAN \cite{park2018data}  employs a Convolutional Neural Network (CNN) for feature processing.
To address these complexities in modeling numerical features, CTGAN \cite{zhao2021ctab} enhances its training procedure by incorporating mode-specific normalization and mitigates data imbalance through the implementation of a conditional generator.

The second group leverage Variational Autoencoder (VAE) methodologies, employing the well-known 'encode-sampling-decode' pipeline. 
VAEM \cite{ma2020vaem} undergoes a two-stage training process. 
A Variational Autoencoder (VAE) is employed to establish a more homogeneous latent representation across dimensions in the first stage. Subsequently, this continuously extracted latent representation is extracted and further utilized in a second VAE for the purpose of dependency modeling.
Building upon the principles introduced by VAEM, TABSYN \cite{zhang2023mixed} incorporates the extracted latent representation into a diffusion model while HH-VAE \cite{peis2022missing} introduces Hamiltonian Monte Carlo to improved approximate inference quality. 
Contrary to the previously mentioned approaches, GOGGLE \cite{liu2022goggle} focuses on learning a graph structure within the latent space.

The third group are constructed based on a diffusion model \cite{}, which approximates the target distribution by examining the endpoint of a Markov chain. This chain originates from a specified parametric distribution, commonly chosen as a standard Gaussian distribution. CoDi \cite{lee2023codi} and TABDDPM \cite{kotelnikov2023tabddpm} are two concurrent works that leverage Gaussian diffusion models and Multinomial diffusion models to formulate numerical and categorical data seperatelly. Current works mainly borrow a similar idea on application-driven tasks, e.g., finance domain \cite{sattarov2023findiff}.
Differently, STaSy \cite{kim2022stasy} proposes a self-paced learning technique and a fine-tuning strategy, which further increases the sampling quality and diversity by stabilizing the denoising score matching training. SOS \cite{kim2022sos} oversamples minor classes since imbalanced classes frequently lead to sub-optimal training
outcomes. 

\subsubsection{Tabular Data Imputation Tasks}
Another line of works address the missing value problems and focus on data imputation.
Predictive approaches to missing data imputation can be categorized in two families \cite{telyatnikov2023egg}: 
(i) imputing missing data through estimating statistics using the entire dataset \cite{lakshminarayan1996imputation,nazabal2020handling};
(ii) inferring the missing components employing similar data points to the one having missing values, e.g, KNN-based method \cite{acuna2004treatment}. To both model feature interaction in a global manner and leverage similar sample information,
recent works \cite{spinelli2020missing} \cite{telyatnikov2023egg} explored the assumption of endowing tabular data
with a graph topology and then exploiting
the message mechanism to impute missing value.
The difficulty and challenges here remain in the definition of a suitable distance metric to compute graph connectivity beforehand and design a customized procedure to sparsify the graph \cite{telyatnikov2023egg}.

\subsubsection{Pretraining on Tabular Data}
\textbf{Challenges} Tabular transference aims to transfer knowledge between tables.
Tabular data exhibit variations in both the number and types of columns (termed as variable-column), posing a challenge for tabular deep learning models to transfer knowledge effectively from one table to another. 
Besides, in contrast to image and text modalities, tabular data are highly domain-specific and often lack extensive, high-quality datasets. 
These three challenges can result in poor generalization abilities across diverse tabular datasets.
Based on the transfering difficulty and data distribution shift types, the core research questions can be classified into the following streads:

\textbf{Sub-Challenge 1: Covariant Shift}. 
Covariate shift is a common scenario in industrial practice, signifying a shift in the marginal distribution of features while the decision boundary of the model, denoted as 
$p(y|x)$, remains unaltered. 
For instance, a patient may undergo multiple visits to a clinical institution, leading to a shifted feature representation. However, the underlying mechanism $p(y|x)$ remains stable.

\textbf{Sub-Challenge 2: Distribution Shift with Varied Columns.} 
In industrial practice, feature design constitutes a critical step where scientists and engineers may introduce new features while removing redundant or unnecessary features. 
Given the substantial overlap in features, retraining the entire model is time-consuming and results in inefficiencies in labor.
However, formulating an effective transfer learning paradigm that incorporates new feature information while discarding removed feature knowledge is a non-trivial task.

\textbf{Sub-Challenge 3: Distribution Shift across Domains}
Vision and text models exhibit adaptability to a diverse array of tasks. This adaptability is attributed to the shared general representations present in both sentences and images, which shows task-agnostic properties \cite{farahani2021brief}. However, in the context of heterogeneous data \cite{ren2022mitigating}, a pertinent question arises: is there shared knowledge across tables, considering that two distinct tables can possess entirely different column numbers and associated semantic meanings?

\textbf{Solution to Sub-Challenge 1 is }within-table pretraining that only covariate shift occurs.
The primary objective is to devise a self-supervised loss and create a corrupted or augmented rendition of the initial tabular data.
In the case of VIME \cite{yoon2020vime}, a mask matrix $\mathcal{M}$ is applied to the initial tabular data $\mathcal{X}$ to generate a corrupted version $\mathcal{\hat{X}}$ as input. The model then undertakes tabular data reconstruction and mask vector estimation, constituting two self-supervised loss components.
Conversely, the subsequent work, SCARF \cite{bahri2021scarf}, generates a corrupted version by replacing each feature with a random draw from its empirical marginal distribution. Additionally, it introduces a contrastive loss.
In contrast to these approaches, SubTab \cite{ucar2021subtab} posits that reconstructing the data from a subset of its features, rather than from its corrupted version in an autoencoder setting, can better capture the underlying latent representation.

\textbf{Solution to Sub-Challenge 2} is across-table pretraining  that formulates a versatile model capable of accommodating variable-column tables. One direction is to convert tabular data (cells in columns) into a sequence of semantically encoded tokens, e.g., TransTab \cite{wang2022transtab}, MED ITAB \cite{wang2023anypredict}, TABRET\cite{onishi2023tabret}, UniTabE \cite{yang2023unitabe}.
Another line of works aim to propose a pseudo-feature method for aligning the upstream and downstream feature sets in heterogeneous data, e.g., \cite{levin2022transfer}.

\textbf{Sub-Solution 3a to Sub-Challenge 3: } is Cross-domain pretraining that goes beyond the limitations imposed by variable-column structures and domain-specific constraints. XTab \cite{zhu2023xtab}, as a pioneering work, engages in pretraining on a diverse set of over 150 tables collected from finance, education, and medical domains. 
To address column variate challenges, Xtab \cite{zhu2023xtab} utilize independent featurizers and use federated learning to pretrain the shared component.
Notably, XTab doesn't strive to learn a universal tokenizer applicable to all tables. Instead, its goal is to acquire a weight initialization that exhibits generalizability across various downstream tasks.

\textbf{Sub-Solution 3b to Sub-Challenge 3}. Currently, transference with LLMs emerges as a new direction for across-domain transference. TabLLM \cite{hegselmann2023tabllm}, Anypredict \cite{wang2023anypredict}, GReaT \cite{borisov2022language} first serialize the feature names and values into a natural language string. This string is then combined with a task-specific prompt for model fine-tuning. Recently, Graphcare \cite{jiang2023graphcare} extracts healthcare-related knowledge graph (KG) from LLMs. The extracted KG is expected to embed generalizable representation and can be used to improve EHR-based predictions.


\begin{table*}
\centering
\begin{tabular}{|l|l|l|}
\hline  

\multirow{3}{*}{Representation} & Homogeneous Representation & \cite{huang2020tabtransformer}, \cite{gorishniy2021revisiting}\\
\cline{2-3}  & Element Representation & \\
\cline{2-3}  & Semantic Representation & \\
\hline 
\multirow{7}{*}{\makecell[l]{Dependency \\ modeling}} & \makecell[l]{Hierarchical \\ structure modeling} & 
\makecell[l]{DeepGBM \cite{ke2019deepgbm}, \cite{popov2019neural},\\ 
GrowNet \cite{badirli2020gradient}, \cite{good2023feature}, \\
TABCAPs \cite{chen2022tabcaps}, TANGOS \cite{jeffares2022tangos}, \\
Net-DNF \cite{katzir2020net}, \cite{zhou2022table2graph}, \\
\cite{yan2023t2g}, \cite{wan2023graphfade}, \\
\cite{liao2023tabgsl}, \cite{du2022learning}}\\
\cline{2-3}  & \makecell[l]{Interactive \\ Structures Modeling} & 
\makecell[l]{TabNet \cite{arik2021tabnet}, \\
SAINT \cite{somepalli2022saint}, NAN \cite{luo2020network}, \\
\cite{silva2020optimization}, \cite{tambwekar2023natural}, \\
\cite{pace2022poetree}, \cite{kalra2023can}} \\
\cline{2-3}  & Latent Structure Discovery & 
\makecell[l]{AGEBO-Tabular \cite{egele2021agebo}, TabNAS \cite{yang2022tabnas}}\\
\hline 

\multirow{6}{*}{Generation} & \makecell[l]{Generative Adversarial \\ Networks} & 
\makecell[l]{medGAN \cite{choi2017generating}, TableGAN \cite{park2018data}, \\
CTGAN \cite{zhao2021ctab}} \\
\cline{2-3}  & Variational Autoencoder & 
\makecell[l]{VAEM \cite{ma2020vaem}, TABSYN \cite{zhang2023mixed}, \\
HH-VAE \cite{peis2022missing}, GOGGLE \cite{liu2022goggle}} \\
\cline{2-3}  & Diffusion Model & 
\makecell[l]{CoDi \cite{lee2023codi}, TABDDPM \cite{kotelnikov2023tabddpm}, \\
\cite{sattarov2023findiff}, STaSy \cite{kim2022stasy}, \\
SOS \cite{kim2022sos}} \\
\hline 

\multirow{7}{*}{Transference} & within table pretraining & 
\makecell[l]{VIME \cite{yoon2020vime}, SCARF \cite{bahri2021scarf}, \\
SubTab \cite{ucar2021subtab}} \\
\cline{2-3}  & across table pretraining & 
\makecell[l]{TransTab \cite{wang2022transtab}, MED ITAB \cite{wang2023anypredict}, \\
TABRET\cite{onishi2023tabret}, UniTabE \cite{yang2023unitabe}, \\
\cite{levin2022transfer}}\\
\cline{2-3}  & across domain pretraining & XTab \cite{zhu2023xtab}\\
\cline{2-3}  & \makecell[l]{transference with \\ Large Language Model} & 
\makecell[l]{TabLLM \cite{hegselmann2023tabllm}, Anypredict \cite{wang2023anypredict}, \\
GReaT \cite{borisov2022language}, Graphcare \cite{jiang2023graphcare},
\\CHAIN-OF-TABLE~\cite{anonymous2024chainoftable}}\\
\hline
\end{tabular}
\caption{\label{tab:widgets}An example table.}
\end{table*}
\section{Conclusion and Future Directions}
\subsection{Representation}
\quad $\clubsuit$ \textbf{Benchmarks}.
Tabular data stands as a prevalent data format in industrial settings; however, its accessibility is frequently constrained due to concerns surrounding data privacy, particularly in domains such as clinical and finance.
The release and synthesis of high-quality tabular data would significantly enhance algorithmic design and contribute substantively to the advancement of representation learning.

$\clubsuit$ \textbf{Theoretical Analysis}. Existing works achieve promising empirical performance on tabular data modeling problems, the theoretical analysis remains an open problem. We believe that rigorous analyses can provide in-depth
insights and inspire the development of new tabular-based methods.
Here we propose several research questions: 
(1) How to formulate and analyze the influence of missing-value problem in tabular representation? 
(2) Different tabular columns generally exhibit a diversity of distribution statistics, e.g., mean, variance. How to evaluate and quantify the significance of such variability in representation learning and loss design?

$\clubsuit$ \textbf{Empirical Analysis} A pioneering work in \cite{} demonstrates that shallow layer in DNN shares generalizable features, e.g., image edge, texture, which builds empirical foundations for transfer learning in image domain. However, empirical validation regarding the transferability and generalization of knowledge within the framework of tabular representations remains largely unexplored.
Further exploration is required to comprehensively investigate and validate these aspects within the tabular domain. 

$\clubsuit$ \textbf{Element Representation}. Existing works mainly consider tabular heterogeneous nature and transform mixed type feature into a unified continuous space. One pioneering work, 
\cite{}, first split numerical features into bins and and find that bin-based embedding could large improve performance. The obtention of element units, especially from the perspective of available external knowledge, is still an under-explored problem in the tabular data domain. 

$\clubsuit$ \textbf{Semantic Representation}. 
Given our primary focus on tabular data featuring numerical and categorical features, there are limited transferable representations within these values. 
However, tabular data is inherently domain-specific, implying that it should exhibit rich domain knowledge. Exploring the usage and interaction with large language models represents a promising direction. This approach would offer auxiliary information, enhancing both the semantic and generalizable representation of tabular data.

\subsection{Dependency Modeling}
\quad $\clubsuit$ \textbf{Temporal Dependency Modeling}
Current studies predominantly address static scenarios wherein tabular features remain constant. Real-world datasets frequently manifest temporal properties, such as sequential Electronic Health Record (EHR) tabular data and sequential financial data. Exploring methodologies for learning dynamic dependencies within temporal tabular data represents a practical and promising direction.

$\clubsuit$ \textbf{Dependency Modeling with Auxiliary Knowledge}
Due to privacy concerns and labor costs, prior research endeavors have explored and modeled tabular data structures without incorporating external knowledge. 
With the advent of Large Language Models (LLMs) and RAG tools \cite{zhang2024ratt}, the prospect of constructing a trustworthy database and knowledge graph to enhance the quality of dependency discovery and the modeling of dependency structures emerges as an intriguing and promising direction.

$\clubsuit$ \textbf{Modeling Data Dependency with Missing Features and Labels} 
In practical settings, tabular data often encompasses instances with missing features or labels. For instance, within Electronic Health Record (EHR) data, certain patient records may exhibit absence of values pertaining to specific medical tests, diagnoses, or other health-related features. The issue of missingness has been underexplored in existing literature.
One important direction would be develop robust dependency models that can effectively handle such scenarios. 
This involves designing algorithms and approaches that can infer dependencies, patterns, or correlations in the presence of missing data, ensuring that the learned dependencies remain applicable and informative even when confronted with incomplete information.

\subsection{Generation} 

\quad $\clubsuit$ \textbf{Sampling quality} 
Current research primarily relies on generative models for the purpose of synthesizing artificial tabular data.
However, realistic tabular data are imbalanced, missingness and heterogeneous. 
A classical generative model exhibits a tendency to emphasize the majority class while overlooking its minority components, resulting in substantial biases affecting the quality of generated data. Furthermore, the presence of missing data introduces significant uncertainty during the generation process. Addressing this challenge involves the incorporation of probabilistic models, representing a crucial direction for improvement.
Besides, Recent work \cite{} has observed that the heterogeneous nature of tabular data causes mode collapse in the latent space when employing VAE-based methods. Considering the heterogeneous property in the latent space and improve sampling quality is an intriguing and important research problem.

$\clubsuit$ \textbf{Structure Modeling}. 
Contemporary research predominantly explores the utilization of advanced generative models for synthesizing tabular data, e.g, from GAN to diffusion models. However, the incorporation of dependency modeling in the generation process is frequently overlooked in recent works.
Leveraging the inherent strengths of data structure modeling, e.g., 
integrate and adapt tree structure into diffusion model stands as an interesting and challenging research question.

\subsection{Transference}
\quad  $\clubsuit$ \textbf{Adaptation without Accessing to Source Data} 
In an industrial setting, retraining models is expensive and time-consuming \cite{ren2022semi}. A more realistic setting considering covariate shift is tabular test time adaptation, where only pretrained model and target tabular feature are available. 
The development of a tailored loss function and adaptation model designed specifically for tabular data holds the potential to yield substantial cost savings for industry companies.

$\clubsuit$ \textbf{Adaptation with Continual Distribution Shift}. 
Current research predominantly focuses on the static nature of tabular data, which does not accurately reflect the dynamic nature inherent in realistic finance or clinical data. The development of a robust and generalizable model capable of handling the evolving distribution shift, e.g., open feature sets, in tabular data would capture the attention of the industry.

$\clubsuit$ \textbf{Adaptation with Large Language Models}. Given that tabular data is inherently domain-specific and may encompass diverse columns, previous works use column imputation strategies to ensure alignment across columns. However, the imputed data often contain significant noise. Exploring the utilization and integration of LLMs for column information imputation emerges as an intriguing and promising direction

\appendix

\bibliographystyle{named}
\bibliography{ijcai23}

\begin{thebibliography}{}

\bibitem[\protect\citeauthoryear{Acuna and Rodriguez}{2004}]{acuna2004treatment}
Edgar Acuna and Caroline Rodriguez.
\newblock The treatment of missing values and its effect on classifier accuracy.
\newblock In {\em Classification, Clustering, and Data Mining Applications: Proceedings of the Meeting of the International Federation of Classification Societies (IFCS), Illinois Institute of Technology, Chicago, 15--18 July 2004}, pages 639--647. Springer, 2004.

\bibitem[\protect\citeauthoryear{Arik and Pfister}{2021}]{arik2021tabnet}
Sercan~{\"O} Arik and Tomas Pfister.
\newblock Tabnet: Attentive interpretable tabular learning.
\newblock In {\em Proceedings of the AAAI conference on artificial intelligence}, volume~35, pages 6679--6687, 2021.

\bibitem[\protect\citeauthoryear{Badirli \bgroup \em et al.\egroup }{2020}]{badirli2020gradient}
Sarkhan Badirli, Xuanqing Liu, Zhengming Xing, Avradeep Bhowmik, Khoa Doan, and Sathiya~S Keerthi.
\newblock Gradient boosting neural networks: Grownet.
\newblock {\em arXiv preprint arXiv:2002.07971}, 2020.

\bibitem[\protect\citeauthoryear{Bahri \bgroup \em et al.\egroup }{2021}]{bahri2021scarf}
Dara Bahri, Heinrich Jiang, Yi~Tay, and Donald Metzler.
\newblock Scarf: Self-supervised contrastive learning using random feature corruption.
\newblock In {\em International Conference on Learning Representations}, 2021.

\bibitem[\protect\citeauthoryear{Borisov \bgroup \em et al.\egroup }{2022a}]{borisov2022deep}
Vadim Borisov, Tobias Leemann, Kathrin Se{\ss}ler, Johannes Haug, Martin Pawelczyk, and Gjergji Kasneci.
\newblock Deep neural networks and tabular data: A survey.
\newblock {\em IEEE transactions on neural networks and learning systems}, 2022.

\bibitem[\protect\citeauthoryear{Borisov \bgroup \em et al.\egroup }{2022b}]{borisov2022language}
Vadim Borisov, Kathrin Se{\ss}ler, Tobias Leemann, Martin Pawelczyk, and Gjergji Kasneci.
\newblock Language models are realistic tabular data generators.
\newblock {\em arXiv preprint arXiv:2210.06280}, 2022.

\bibitem[\protect\citeauthoryear{Chen and Guestrin}{2016}]{chen2016xgboost}
Tianqi Chen and Carlos Guestrin.
\newblock Xgboost: A scalable tree boosting system.
\newblock In {\em Proceedings of the 22nd acm sigkdd international conference on knowledge discovery and data mining}, pages 785--794, 2016.

\bibitem[\protect\citeauthoryear{Chen \bgroup \em et al.\egroup }{2022}]{chen2022tabcaps}
Jintai Chen, KuanLun Liao, Yanwen Fang, Danny Chen, and Jian Wu.
\newblock Tabcaps: A capsule neural network for tabular data classification with bow routing.
\newblock In {\em The Eleventh International Conference on Learning Representations}, 2022.

\bibitem[\protect\citeauthoryear{Choi \bgroup \em et al.\egroup }{2017}]{choi2017generating}
Edward Choi, Siddharth Biswal, Bradley Malin, Jon Duke, Walter~F Stewart, and Jimeng Sun.
\newblock Generating multi-label discrete patient records using generative adversarial networks.
\newblock In {\em Machine learning for healthcare conference}, pages 286--305. PMLR, 2017.

\bibitem[\protect\citeauthoryear{Du \bgroup \em et al.\egroup }{2022}]{du2022learning}
Kounianhua Du, Weinan Zhang, Ruiwen Zhou, Yangkun Wang, Xilong Zhao, Jiarui Jin, Quan Gan, Zheng Zhang, and David~P Wipf.
\newblock Learning enhanced representation for tabular data via neighborhood propagation.
\newblock {\em Advances in Neural Information Processing Systems}, 35:16373--16384, 2022.

\bibitem[\protect\citeauthoryear{Egele \bgroup \em et al.\egroup }{2021}]{egele2021agebo}
Romain Egele, Prasanna Balaprakash, Isabelle Guyon, Venkatram Vishwanath, Fangfang Xia, Rick Stevens, and Zhengying Liu.
\newblock Agebo-tabular: joint neural architecture and hyperparameter search with autotuned data-parallel training for tabular data.
\newblock In {\em Proceedings of the International Conference for High Performance Computing, Networking, Storage and Analysis}, pages 1--14, 2021.

\bibitem[\protect\citeauthoryear{Farahani \bgroup \em et al.\egroup }{2021}]{farahani2021brief}
Abolfazl Farahani, Sahar Voghoei, Khaled Rasheed, and Hamid~R Arabnia.
\newblock A brief review of domain adaptation.
\newblock {\em Advances in data science and information engineering: proceedings from ICDATA 2020 and IKE 2020}, pages 877--894, 2021.

\bibitem[\protect\citeauthoryear{Fern{\'a}ndez-Delgado \bgroup \em et al.\egroup }{2014}]{fernandez2014we}
Manuel Fern{\'a}ndez-Delgado, Eva Cernadas, Sen{\'e}n Barro, and Dinani Amorim.
\newblock Do we need hundreds of classifiers to solve real world classification problems?
\newblock {\em The journal of machine learning research}, 15(1):3133--3181, 2014.

\bibitem[\protect\citeauthoryear{Good \bgroup \em et al.\egroup }{2023}]{good2023feature}
Jack~Henry Good, Torin Kovach, Kyle Miller, and Artur Dubrawski.
\newblock Feature learning for interpretable, performant decision trees.
\newblock In {\em Thirty-seventh Conference on Neural Information Processing Systems}, 2023.

\bibitem[\protect\citeauthoryear{Goodfellow \bgroup \em et al.\egroup }{2020}]{goodfellow2020generative}
Ian Goodfellow, Jean Pouget-Abadie, Mehdi Mirza, Bing Xu, David Warde-Farley, Sherjil Ozair, Aaron Courville, and Yoshua Bengio.
\newblock Generative adversarial networks.
\newblock {\em Communications of the ACM}, 63(11):139--144, 2020.

\bibitem[\protect\citeauthoryear{Gorishniy \bgroup \em et al.\egroup }{2021}]{gorishniy2021revisiting}
Yury Gorishniy, Ivan Rubachev, Valentin Khrulkov, and Artem Babenko.
\newblock Revisiting deep learning models for tabular data.
\newblock {\em Advances in Neural Information Processing Systems}, 34:18932--18943, 2021.

\bibitem[\protect\citeauthoryear{Hegselmann \bgroup \em et al.\egroup }{2023}]{hegselmann2023tabllm}
Stefan Hegselmann, Alejandro Buendia, Hunter Lang, Monica Agrawal, Xiaoyi Jiang, and David Sontag.
\newblock Tabllm: Few-shot classification of tabular data with large language models.
\newblock In {\em International Conference on Artificial Intelligence and Statistics}, pages 5549--5581. PMLR, 2023.

\bibitem[\protect\citeauthoryear{Huang \bgroup \em et al.\egroup }{2020}]{huang2020tabtransformer}
Xin Huang, Ashish Khetan, Milan Cvitkovic, and Zohar Karnin.
\newblock Tabtransformer: Tabular data modeling using contextual embeddings.
\newblock {\em arXiv preprint arXiv:2012.06678}, 2020.

\bibitem[\protect\citeauthoryear{Jeffares \bgroup \em et al.\egroup }{2022}]{jeffares2022tangos}
Alan Jeffares, Tennison Liu, Jonathan Crabb{\'e}, Fergus Imrie, and Mihaela van~der Schaar.
\newblock Tangos: Regularizing tabular neural networks through gradient orthogonalization and specialization.
\newblock In {\em The Eleventh International Conference on Learning Representations}, 2022.

\bibitem[\protect\citeauthoryear{Jiang \bgroup \em et al.\egroup }{2023}]{jiang2023graphcare}
Pengcheng Jiang, Cao Xiao, Adam Cross, and Jimeng Sun.
\newblock Graphcare: Enhancing healthcare predictions with open-world personalized knowledge graphs.
\newblock {\em arXiv preprint arXiv:2305.12788}, 2023.

\bibitem[\protect\citeauthoryear{Kalra and Brown}{2023}]{kalra2023can}
Akansha Kalra and Daniel~S Brown.
\newblock Can differentiable decision trees learn interpretable reward functions?
\newblock {\em arXiv preprint arXiv:2306.13004}, 2023.

\bibitem[\protect\citeauthoryear{Katzir \bgroup \em et al.\egroup }{2020}]{katzir2020net}
Liran Katzir, Gal Elidan, and Ran El-Yaniv.
\newblock Net-dnf: Effective deep modeling of tabular data.
\newblock In {\em International conference on learning representations}, 2020.

\bibitem[\protect\citeauthoryear{Ke \bgroup \em et al.\egroup }{2019}]{ke2019deepgbm}
Guolin Ke, Zhenhui Xu, Jia Zhang, Jiang Bian, and Tie-Yan Liu.
\newblock Deepgbm: A deep learning framework distilled by gbdt for online prediction tasks.
\newblock In {\em Proceedings of the 25th ACM SIGKDD International Conference on Knowledge Discovery \& Data Mining}, pages 384--394, 2019.

\bibitem[\protect\citeauthoryear{Kim \bgroup \em et al.\egroup }{2022a}]{kim2022stasy}
Jayoung Kim, Chaejeong Lee, and Noseong Park.
\newblock Stasy: Score-based tabular data synthesis.
\newblock In {\em The Eleventh International Conference on Learning Representations}, 2022.

\bibitem[\protect\citeauthoryear{Kim \bgroup \em et al.\egroup }{2022b}]{kim2022sos}
Jayoung Kim, Chaejeong Lee, Yehjin Shin, Sewon Park, Minjung Kim, Noseong Park, and Jihoon Cho.
\newblock Sos: Score-based oversampling for tabular data.
\newblock In {\em Proceedings of the 28th ACM SIGKDD Conference on Knowledge Discovery and Data Mining}, pages 762--772, 2022.

\bibitem[\protect\citeauthoryear{Kingma}{2013}]{kingma2013auto}
Diederik~P Kingma.
\newblock Auto-encoding variational bayes.
\newblock {\em arXiv preprint arXiv:1312.6114}, 2013.

\bibitem[\protect\citeauthoryear{Kotelnikov \bgroup \em et al.\egroup }{2023}]{kotelnikov2023tabddpm}
Akim Kotelnikov, Dmitry Baranchuk, Ivan Rubachev, and Artem Babenko.
\newblock Tabddpm: Modelling tabular data with diffusion models.
\newblock In {\em International Conference on Machine Learning}, pages 17564--17579. PMLR, 2023.

\bibitem[\protect\citeauthoryear{Lakshminarayan \bgroup \em et al.\egroup }{1996}]{lakshminarayan1996imputation}
Kamakshi Lakshminarayan, Steven~A Harp, Robert~P Goldman, Tariq Samad, et~al.
\newblock Imputation of missing data using machine learning techniques.
\newblock In {\em KDD}, volume~96, 1996.

\bibitem[\protect\citeauthoryear{Lee \bgroup \em et al.\egroup }{2023}]{lee2023codi}
Chaejeong Lee, Jayoung Kim, and Noseong Park.
\newblock Codi: Co-evolving contrastive diffusion models for mixed-type tabular synthesis.
\newblock {\em arXiv preprint arXiv:2304.12654}, 2023.

\bibitem[\protect\citeauthoryear{Levin \bgroup \em et al.\egroup }{2022}]{levin2022transfer}
Roman Levin, Valeriia Cherepanova, Avi Schwarzschild, Arpit Bansal, C~Bayan Bruss, Tom Goldstein, Andrew~Gordon Wilson, and Micah Goldblum.
\newblock Transfer learning with deep tabular models.
\newblock {\em arXiv preprint arXiv:2206.15306}, 2022.

\bibitem[\protect\citeauthoryear{Li \bgroup \em et al.\egroup }{2019}]{li2019lumen}
Ziyan Li, Jianjiang Feng, Zishun Feng, Yunqiang An, Yang Gao, Bin Lu, and Jie Zhou.
\newblock Lumen segmentation of aortic dissection with cascaded convolutional network.
\newblock In {\em Statistical Atlases and Computational Models of the Heart. Atrial Segmentation and LV Quantification Challenges: 9th International Workshop, STACOM 2018, Held in Conjunction with MICCAI 2018, Granada, Spain, September 16, 2018, Revised Selected Papers 9}, pages 122--130. Springer, 2019.

\bibitem[\protect\citeauthoryear{Liang \bgroup \em et al.\egroup }{2024}]{liang2024inducing}
Junjie Liang, Weijieying Ren, Hanifi Sahar, and Vasant Honavar.
\newblock Inducing clusters deep kernel gaussian process for longitudinal data.
\newblock In {\em Proceedings of the AAAI Conference on Artificial Intelligence}, volume~38, pages 13736--13743, 2024.

\bibitem[\protect\citeauthoryear{Liao and Li}{2023}]{liao2023tabgsl}
Jay~Chiehen Liao and Cheng-Te Li.
\newblock Tabgsl: Graph structure learning for tabular data prediction.
\newblock {\em arXiv preprint arXiv:2305.15843}, 2023.

\bibitem[\protect\citeauthoryear{Liu \bgroup \em et al.\egroup }{2017}]{liu2017left}
Honghui Liu, Jianjiang Feng, Zishun Feng, Jiwen Lu, and Jie Zhou.
\newblock Left atrium segmentation in ct volumes with fully convolutional networks.
\newblock In {\em Deep Learning in Medical Image Analysis and Multimodal Learning for Clinical Decision Support: Third International Workshop, DLMIA 2017, and 7th International Workshop, ML-CDS 2017, Held in Conjunction with MICCAI 2017, Qu{\'e}bec City, QC, Canada, September 14, Proceedings 3}, pages 39--46. Springer, 2017.

\bibitem[\protect\citeauthoryear{Liu \bgroup \em et al.\egroup }{2022}]{liu2022goggle}
Tennison Liu, Zhaozhi Qian, Jeroen Berrevoets, and Mihaela van~der Schaar.
\newblock Goggle: Generative modelling for tabular data by learning relational structure.
\newblock In {\em The Eleventh International Conference on Learning Representations}, 2022.

\bibitem[\protect\citeauthoryear{Liu \bgroup \em et al.\egroup }{2023}]{liu2023tabular}
Jixiong Liu, Yoan Chabot, Rapha{\"e}l Troncy, Viet-Phi Huynh, Thomas Labb{\'e}, and Pierre Monnin.
\newblock From tabular data to knowledge graphs: A survey of semantic table interpretation tasks and methods.
\newblock {\em Journal of Web Semantics}, 76:100761, 2023.

\bibitem[\protect\citeauthoryear{Luo \bgroup \em et al.\egroup }{2020}]{luo2020network}
Yuanfei Luo, Hao Zhou, Wei-Wei Tu, Yuqiang Chen, Wenyuan Dai, and Qiang Yang.
\newblock Network on network for tabular data classification in real-world applications.
\newblock In {\em Proceedings of the 43rd International ACM SIGIR Conference on Research and Development in Information Retrieval}, pages 2317--2326, 2020.

\bibitem[\protect\citeauthoryear{Ma \bgroup \em et al.\egroup }{2020}]{ma2020vaem}
Chao Ma, Sebastian Tschiatschek, Richard Turner, Jos{\'e}~Miguel Hern{\'a}ndez-Lobato, and Cheng Zhang.
\newblock Vaem: a deep generative model for heterogeneous mixed type data.
\newblock {\em Advances in Neural Information Processing Systems}, 33:11237--11247, 2020.

\bibitem[\protect\citeauthoryear{Ma \bgroup \em et al.\egroup }{2022a}]{ma2022elucidating}
Haixu Ma, Yufeng Liu, and Guorong Wu.
\newblock Elucidating multi-stage progression of neuro-degeneration process in alzheimer’s disease.
\newblock {\em Alzheimer's \& Dementia}, 18:e068774, 2022.

\bibitem[\protect\citeauthoryear{Ma \bgroup \em et al.\egroup }{2022b}]{ma2022learning}
Haixu Ma, Donglin Zeng, and Yufeng Liu.
\newblock Learning individualized treatment rules with many treatments: A supervised clustering approach using adaptive fusion.
\newblock {\em Advances in Neural Information Processing Systems}, 35:15956--15969, 2022.

\bibitem[\protect\citeauthoryear{Ma \bgroup \em et al.\egroup }{2023}]{ma2023learning}
Haixu Ma, Donglin Zeng, and Yufeng Liu.
\newblock Learning optimal group-structured individualized treatment rules with many treatments.
\newblock {\em Journal of Machine Learning Research}, 24(102):1--48, 2023.

\bibitem[\protect\citeauthoryear{Mackinnon}{2010}]{mackinnon2010use}
A~Mackinnon.
\newblock The use and reporting of multiple imputation in medical research--a review.
\newblock {\em Journal of internal medicine}, 268(6):586--593, 2010.

\bibitem[\protect\citeauthoryear{Nazabal \bgroup \em et al.\egroup }{2020}]{nazabal2020handling}
Alfredo Nazabal, Pablo~M Olmos, Zoubin Ghahramani, and Isabel Valera.
\newblock Handling incomplete heterogeneous data using vaes.
\newblock {\em Pattern Recognition}, 107:107501, 2020.

\bibitem[\protect\citeauthoryear{Nguyen \bgroup \em et al.\egroup }{2019}]{nguyen2019mtab}
Phuc Nguyen, Natthawut Kertkeidkachorn, Ryutaro Ichise, and Hideaki Takeda.
\newblock Mtab: Matching tabular data to knowledge graph using probability models.
\newblock {\em arXiv preprint arXiv:1910.00246}, 2019.

\bibitem[\protect\citeauthoryear{Onishi \bgroup \em et al.\egroup }{2023}]{onishi2023tabret}
Soma Onishi, Kenta Oono, and Kohei Hayashi.
\newblock Tabret: Pre-training transformer-based tabular models for unseen columns.
\newblock {\em arXiv preprint arXiv:2303.15747}, 2023.

\bibitem[\protect\citeauthoryear{Pace \bgroup \em et al.\egroup }{2022}]{pace2022poetree}
Aliz{\'e}e Pace, Alex~J Chan, and Mihaela van~der Schaar.
\newblock Poetree: Interpretable policy learning with adaptive decision trees.
\newblock {\em arXiv preprint arXiv:2203.08057}, 2022.

\bibitem[\protect\citeauthoryear{Park \bgroup \em et al.\egroup }{2018}]{park2018data}
Noseong Park, Mahmoud Mohammadi, Kshitij Gorde, Sushil Jajodia, Hongkyu Park, and Youngmin Kim.
\newblock Data synthesis based on generative adversarial networks.
\newblock {\em arXiv preprint arXiv:1806.03384}, 2018.

\bibitem[\protect\citeauthoryear{Peis \bgroup \em et al.\egroup }{2022}]{peis2022missing}
Ignacio Peis, Chao Ma, and Jos{\'e}~Miguel Hern{\'a}ndez-Lobato.
\newblock Missing data imputation and acquisition with deep hierarchical models and hamiltonian monte carlo.
\newblock {\em Advances in Neural Information Processing Systems}, 35:35839--35851, 2022.

\bibitem[\protect\citeauthoryear{Popov \bgroup \em et al.\egroup }{2019}]{popov2019neural}
Sergei Popov, Stanislav Morozov, and Artem Babenko.
\newblock Neural oblivious decision ensembles for deep learning on tabular data.
\newblock In {\em International Conference on Learning Representations}, 2019.

\bibitem[\protect\citeauthoryear{Qin \bgroup \em et al.\egroup }{2023}]{qin2023read}
Wei Qin, Zetong Chen, Lei Wang, Yunshi Lan, Weijieying Ren, and Richang Hong.
\newblock Read, diagnose and chat: Towards explainable and interactive llms-augmented depression detection in social media.
\newblock {\em arXiv preprint arXiv:2305.05138}, 2023.

\bibitem[\protect\citeauthoryear{Ren and Honavar}{2024}]{ren2024esacl}
Weijieying Ren and Vasant~G Honavar.
\newblock Esacl: An efficient continual learning algorithm.
\newblock In {\em Proceedings of the 2024 SIAM International Conference on Data Mining (SDM)}, pages 163--171. SIAM, 2024.

\bibitem[\protect\citeauthoryear{Ren \bgroup \em et al.\egroup }{2022a}]{ren2022mitigating}
Weijieying Ren, Lei Wang, Kunpeng Liu, Ruocheng Guo, Lim~Ee Peng, and Yanjie Fu.
\newblock Mitigating popularity bias in recommendation with unbalanced interactions: A gradient perspective.
\newblock In {\em 2022 IEEE International Conference on Data Mining (ICDM)}, pages 438--447. IEEE, 2022.

\bibitem[\protect\citeauthoryear{Ren \bgroup \em et al.\egroup }{2022b}]{ren2022semi}
Weijieying Ren, Pengyang Wang, Xiaolin Li, Charles~E Hughes, and Yanjie Fu.
\newblock Semi-supervised drifted stream learning with short lookback.
\newblock In {\em Proceedings of the 28th ACM SIGKDD Conference on Knowledge Discovery and Data Mining}, pages 1504--1513, 2022.

\bibitem[\protect\citeauthoryear{Ren \bgroup \em et al.\egroup }{2024}]{rentablog}
Weijieying Ren, Xiaoting Li, Huiyuan Chen, Vineeth Rakesh, Zhuoyi Wang, Mahashweta Das, and Vasant~G Honavar.
\newblock Tablog: Test-time adaptation for tabular data using logic rules.
\newblock In {\em Forty-first International Conference on Machine Learning}, 2024.

\bibitem[\protect\citeauthoryear{Ruan \bgroup \em et al.\egroup }{2024}]{ruan2024language}
Yucheng Ruan, Xiang Lan, Jingying Ma, Yizhi Dong, Kai He, and Mengling Feng.
\newblock Language modeling on tabular data: A survey of foundations, techniques and evolution.
\newblock {\em arXiv preprint arXiv:2408.10548}, 2024.

\bibitem[\protect\citeauthoryear{Ruiz \bgroup \em et al.\egroup }{2023}]{ruiz2023high}
Camilo Ruiz, Hongyu Ren, Kexin Huang, and Jure Leskovec.
\newblock High dimensional, tabular deep learning with an auxiliary knowledge graph.
\newblock In {\em Thirty-seventh Conference on Neural Information Processing Systems}, 2023.

\bibitem[\protect\citeauthoryear{Sahakyan \bgroup \em et al.\egroup }{2021}]{sahakyan2021explainable}
Maria Sahakyan, Zeyar Aung, and Talal Rahwan.
\newblock Explainable artificial intelligence for tabular data: A survey.
\newblock {\em IEEE access}, 9:135392--135422, 2021.

\bibitem[\protect\citeauthoryear{Santos \bgroup \em et al.\egroup }{2022}]{santos2022knowledge}
Alberto Santos, Ana~R Cola{\c{c}}o, Annelaura~B Nielsen, Lili Niu, Maximilian Strauss, Philipp~E Geyer, Fabian Coscia, Nicolai J~Wewer Albrechtsen, Filip Mundt, Lars~Juhl Jensen, et~al.
\newblock A knowledge graph to interpret clinical proteomics data.
\newblock {\em Nature biotechnology}, 40(5):692--702, 2022.

\bibitem[\protect\citeauthoryear{Sattarov \bgroup \em et al.\egroup }{2023}]{sattarov2023findiff}
Timur Sattarov, Marco Schreyer, and Damian Borth.
\newblock Findiff: Diffusion models for financial tabular data generation.
\newblock In {\em Proceedings of the Fourth ACM International Conference on AI in Finance}, pages 64--72, 2023.

\bibitem[\protect\citeauthoryear{Sauber-Cole and Khoshgoftaar}{2022}]{sauber2022use}
Rick Sauber-Cole and Taghi~M Khoshgoftaar.
\newblock The use of generative adversarial networks to alleviate class imbalance in tabular data: a survey.
\newblock {\em Journal of Big Data}, 9(1):98, 2022.

\bibitem[\protect\citeauthoryear{Silva \bgroup \em et al.\egroup }{2020}]{silva2020optimization}
Andrew Silva, Matthew Gombolay, Taylor Killian, Ivan Jimenez, and Sung-Hyun Son.
\newblock Optimization methods for interpretable differentiable decision trees applied to reinforcement learning.
\newblock In {\em International conference on artificial intelligence and statistics}, pages 1855--1865. PMLR, 2020.

\bibitem[\protect\citeauthoryear{Somepalli \bgroup \em et al.\egroup }{2022}]{somepalli2022saint}
Gowthami Somepalli, Avi Schwarzschild, Micah Goldblum, C~Bayan Bruss, and Tom Goldstein.
\newblock Saint: Improved neural networks for tabular data via row attention and contrastive pre-training.
\newblock In {\em NeurIPS 2022 First Table Representation Workshop}, 2022.

\bibitem[\protect\citeauthoryear{Spinelli \bgroup \em et al.\egroup }{2020}]{spinelli2020missing}
Indro Spinelli, Simone Scardapane, and Aurelio Uncini.
\newblock Missing data imputation with adversarially-trained graph convolutional networks.
\newblock {\em Neural Networks}, 129:249--260, 2020.

\bibitem[\protect\citeauthoryear{Tambwekar \bgroup \em et al.\egroup }{2023}]{tambwekar2023natural}
Pradyumna Tambwekar, Andrew Silva, Nakul Gopalan, and Matthew Gombolay.
\newblock Natural language specification of reinforcement learning policies through differentiable decision trees.
\newblock {\em IEEE Robotics and Automation Letters}, 2023.

\bibitem[\protect\citeauthoryear{Telyatnikov and Scardapane}{2023}]{telyatnikov2023egg}
Lev Telyatnikov and Simone Scardapane.
\newblock Egg-gae: scalable graph neural networks for tabular data imputation.
\newblock In {\em International Conference on Artificial Intelligence and Statistics}, pages 2661--2676. PMLR, 2023.

\bibitem[\protect\citeauthoryear{Ucar \bgroup \em et al.\egroup }{2021}]{ucar2021subtab}
Talip Ucar, Ehsan Hajiramezanali, and Lindsay Edwards.
\newblock Subtab: Subsetting features of tabular data for self-supervised representation learning.
\newblock {\em Advances in Neural Information Processing Systems}, 34:18853--18865, 2021.

\bibitem[\protect\citeauthoryear{Wan \bgroup \em et al.\egroup }{2023}]{wan2023graphfade}
Junhong Wan, Yao Fu, Junlan Yu, Weihao Jiang, Shiliang Pu, and Ruiheng Yang.
\newblock Graphfade: Field-aware decorrelation neural network for graphs with tabular features.
\newblock In {\em Proceedings of the 32nd ACM International Conference on Information and Knowledge Management}, pages 2502--2511, 2023.

\bibitem[\protect\citeauthoryear{Wang and Sun}{2022}]{wang2022transtab}
Zifeng Wang and Jimeng Sun.
\newblock Transtab: Learning transferable tabular transformers across tables.
\newblock {\em Advances in Neural Information Processing Systems}, 35:2902--2915, 2022.

\bibitem[\protect\citeauthoryear{Wang \bgroup \em et al.\egroup }{2023}]{wang2023anypredict}
Zifeng Wang, Chufan Gao, Cao Xiao, and Jimeng Sun.
\newblock Anypredict: Foundation model for tabular prediction.
\newblock {\em arXiv preprint arXiv:2305.12081}, 2023.

\bibitem[\protect\citeauthoryear{Wang \bgroup \em et al.\egroup }{2024a}]{wang2024survey}
Wei-Yao Wang, Wei-Wei Du, Derek Xu, Wei Wang, and Wen-Chih Peng.
\newblock A survey on self-supervised learning for non-sequential tabular data.
\newblock {\em arXiv preprint arXiv:2402.01204}, 2024.

\bibitem[\protect\citeauthoryear{Wang \bgroup \em et al.\egroup }{2024b}]{anonymous2024chainoftable}
Zilong Wang, Hao Zhang, Chun-Liang Li, Julian~Martin Eisenschlos, Vincent Perot, Zifeng Wang, Lesly Miculicich, Yasuhisa Fujii, Jingbo Shang, Chen-Yu Lee, et~al.
\newblock Chain-of-table: Evolving tables in the reasoning chain for table understanding.
\newblock In {\em The Twelfth International Conference on Learning Representations}, 2024.

\bibitem[\protect\citeauthoryear{Wijmans and Baker}{1995}]{wijmans1995solution}
Johannes~G Wijmans and Richard~W Baker.
\newblock The solution-diffusion model: a review.
\newblock {\em Journal of membrane science}, 107(1-2):1--21, 1995.

\bibitem[\protect\citeauthoryear{Yan \bgroup \em et al.\egroup }{2023}]{yan2023t2g}
Jiahuan Yan, Jintai Chen, Yixuan Wu, Danny~Z Chen, and Jian Wu.
\newblock T2g-former: organizing tabular features into relation graphs promotes heterogeneous feature interaction.
\newblock In {\em Proceedings of the AAAI Conference on Artificial Intelligence}, volume~37, pages 10720--10728, 2023.

\bibitem[\protect\citeauthoryear{Yang \bgroup \em et al.\egroup }{2022}]{yang2022tabnas}
Chengrun Yang, Gabriel Bender, Hanxiao Liu, Pieter-Jan Kindermans, Madeleine Udell, Yifeng Lu, Quoc~V Le, and Da~Huang.
\newblock Tabnas: Rejection sampling for neural architecture search on tabular datasets.
\newblock {\em Advances in Neural Information Processing Systems}, 35:11906--11917, 2022.

\bibitem[\protect\citeauthoryear{Yang \bgroup \em et al.\egroup }{2023}]{yang2023unitabe}
Yazheng Yang, Yuqi Wang, Guang Liu, Ledell Wu, and Qi~Liu.
\newblock Unitabe: Pretraining a unified tabular encoder for heterogeneous tabular data.
\newblock {\em arXiv preprint arXiv:2307.09249}, 2023.

\bibitem[\protect\citeauthoryear{Yoon \bgroup \em et al.\egroup }{2020}]{yoon2020vime}
Jinsung Yoon, Yao Zhang, James Jordon, and Mihaela van~der Schaar.
\newblock Vime: Extending the success of self-and semi-supervised learning to tabular domain.
\newblock {\em Advances in Neural Information Processing Systems}, 33:11033--11043, 2020.

\bibitem[\protect\citeauthoryear{Zhang \bgroup \em et al.\egroup }{2023}]{zhang2023mixed}
Hengrui Zhang, Jiani Zhang, Balasubramaniam Srinivasan, Zhengyuan Shen, Xiao Qin, Christos Faloutsos, Huzefa Rangwala, and George Karypis.
\newblock Mixed-type tabular data synthesis with score-based diffusion in latent space.
\newblock {\em arXiv preprint arXiv:2310.09656}, 2023.

\bibitem[\protect\citeauthoryear{Zhang \bgroup \em et al.\egroup }{2024}]{zhang2024ratt}
Jinghan Zhang, Xiting Wang, Weijieying Ren, Lu~Jiang, Dongjie Wang, and Kunpeng Liu.
\newblock Ratt: Athought structure for coherent and correct llmreasoning.
\newblock {\em arXiv preprint arXiv:2406.02746}, 2024.

\bibitem[\protect\citeauthoryear{Zhao \bgroup \em et al.\egroup }{2021}]{zhao2021ctab}
Zilong Zhao, Aditya Kunar, Robert Birke, and Lydia~Y Chen.
\newblock Ctab-gan: Effective table data synthesizing.
\newblock In {\em Asian Conference on Machine Learning}, pages 97--112. PMLR, 2021.

\bibitem[\protect\citeauthoryear{Zhao \bgroup \em et al.\egroup }{2022a}]{zhao2022towards}
Tianxiang Zhao, Enyan Dai, Kai Shu, and Suhang Wang.
\newblock Towards fair classifiers without sensitive attributes: Exploring biases in related features.
\newblock In {\em Proceedings of the Fifteenth ACM International Conference on Web Search and Data Mining}, pages 1433--1442, 2022.

\bibitem[\protect\citeauthoryear{Zhao \bgroup \em et al.\egroup }{2022b}]{zhao2022exploring}
Tianxiang Zhao, Xiang Zhang, and Suhang Wang.
\newblock Exploring edge disentanglement for node classification.
\newblock In {\em Proceedings of the ACM Web Conference 2022}, pages 1028--1036, 2022.

\bibitem[\protect\citeauthoryear{Zhao \bgroup \em et al.\egroup }{2023}]{zhao2023skill}
Tianxiang Zhao, Wenchao Yu, Suhang Wang, Lu~Wang, Xiang Zhang, Yuncong Chen, Yanchi Liu, Wei Cheng, and Haifeng Chen.
\newblock Skill disentanglement for imitation learning from suboptimal demonstrations.
\newblock In {\em Proceedings of the 29th ACM SIGKDD Conference on Knowledge Discovery and Data Mining}, pages 3513--3524, 2023.

\bibitem[\protect\citeauthoryear{Zhao \bgroup \em et al.\egroup }{2024}]{zhao2024interpretable}
Tianxiang Zhao, Wenchao Yu, Suhang Wang, Lu~Wang, Xiang Zhang, Yuncong Chen, Yanchi Liu, Wei Cheng, and Haifeng Chen.
\newblock Interpretable imitation learning with dynamic causal relations.
\newblock In {\em Proceedings of the 17th ACM International Conference on Web Search and Data Mining}, pages 967--975, 2024.

\bibitem[\protect\citeauthoryear{Zhou \bgroup \em et al.\egroup }{2022}]{zhou2022table2graph}
Kaixiong Zhou, Zirui Liu, Rui Chen, Li~Li, S~Choi, and Xia Hu.
\newblock Table2graph: Transforming tabular data to unified weighted graph.
\newblock In {\em Proceedings of the Thirty-First International Joint Conference on Artificial Intelligence, IJCAI}, pages 2420--2426, 2022.

\bibitem[\protect\citeauthoryear{Zhu \bgroup \em et al.\egroup }{2020}]{zhu2020beyond}
Jiong Zhu, Yujun Yan, Lingxiao Zhao, Mark Heimann, Leman Akoglu, and Danai Koutra.
\newblock Beyond homophily in graph neural networks: Current limitations and effective designs.
\newblock {\em Advances in neural information processing systems}, 33:7793--7804, 2020.

\bibitem[\protect\citeauthoryear{Zhu \bgroup \em et al.\egroup }{2023}]{zhu2023xtab}
Bingzhao Zhu, Xingjian Shi, Nick Erickson, Mu~Li, George Karypis, and Mahsa Shoaran.
\newblock Xtab: Cross-table pretraining for tabular transformers.
\newblock {\em arXiv preprint arXiv:2305.06090}, 2023.

\end{thebibliography}

\end{document}